\def\year{2019}\relax
\documentclass[letterpaper]{article} 
\usepackage{aaai19}  
\usepackage{times}  
\usepackage{helvet} 
\usepackage{courier}  
\usepackage[hyphens]{url}  
\usepackage{graphicx} 
\def\UrlFont{\rm}  
\usepackage{graphicx}  
\usepackage{xspace}
\usepackage{xcolor}
\usepackage{mathtools}
\usepackage{amssymb}
\usepackage{subcaption}
\usepackage{booktabs}

\newcommand{\dba}{\texttt{D3BA}\xspace}
\newcommand{\dwa}{\texttt{D3WA}\xspace}

\nocopyright

\title{\dba: A Tool for Optimizing Business Processes\\Using
Non-Deterministic Planning}
\author{
Tathagata Chakraborti \and Yasaman Khazaeni \\
IBM Research AI, Cambridge MA USA\\[1ex]
{\em Contact:} {\tt tchakra2@ibm.com}, {\tt yasaman.khazaeni@us.ibm.com}\\[2ex]
\textnormal{\em Business Process + Skills = Optimized Business Process}
}

\begin{document}

\maketitle

\begin{abstract}
This paper builds upon recent work in the declarative design
of dialogue agents and proposes an exciting new tool -- \dba\ -- Declarative Design for Digital Business Automation, built to optimize business processes using the power of AI planning.
The tool provides a powerful framework  
to build, optimize, and maintain complex business processes
and optimize them by composing with services
that automate one or more subtasks. 
We illustrate salient features of this composition technique,
compare with other philosophies of composition,
and highlight exciting opportunities for research in this
emerging field of business process automation.
\end{abstract}

\begin{figure}
\centering
\includegraphics[width=\columnwidth]{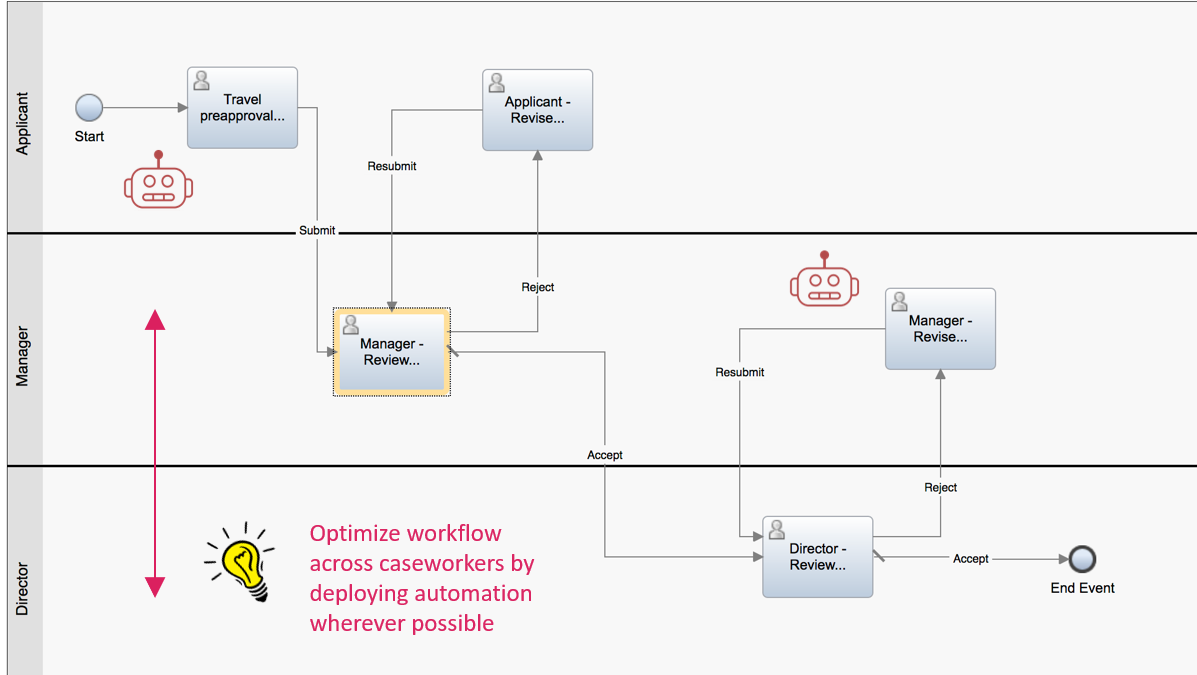}
\caption{A snapshot of a simple travel authorization application
from IBM's Business Process Management tool. 
The different layers represent handovers between caseworkers
with different privileges. The boxes representing individual
subtasks are connected through logic gates (AND, OR, etc.) 
that control the overall workflow of the process.}
\label{fig:travel}
\end{figure}

\noindent A business process is a collection of tasks which in a specific sequence meet some business goal, such as produce a service or product for customers. People performing these tasks are referred to as case workers.
Figure~\ref{fig:travel} illustrates a simple business process 
involving the approval of a trip request from an employee to present at a conference. The process begins with a stage of information gathering from the employee, followed by reviews by their manager and director, ending with eventual acceptance or rejection. 

In practice, the life cycle of a business process is riddled with repetitive tasks, severe bottlenecks and hot spots which impact the performance of the case worker, and quality of service.  
Recent advances in artificial intelligence can be leveraged to significantly revamp how we build and maintain business processes with the goal of improving the case worker experience. 
Indeed the ability to inject manual business processes with 
artificial intelligence and sophisticated automation
has received increased attention lately \cite{hull2016rethinking}. 
For example, in the travel approval process described above, there are stages where automation can be used to speed up the workflow significantly, such as during the acquisition of information for the applicant. 
We want to be able to determine how best to do this, 
where and when to deploy automation, 
to maximize impact on the process and reduce load on the individual caseworkers.

\section{Business Processes and Planning}

Business process management has been one of the most actively
researched areas when it comes to applications of planning technologies. 
A fantastic guide to existing work and challenges at the 
intersection of planning and business process management
can be found in \cite{marrella2017automated}.
Salient problems addressed in this area include specification
and construction of business processes in the form of planning problems \cite{r2007integrating},
robustification and adaptation to failures \cite{jarvis1999exploiting},
validation, verification, and monitoring of processes \cite{de2018aligning}, 
and prediction and mitigation of risk \cite{sohrabi2018ai}.
We address a somewhat overlooked promise of 
this union -- that of using planning to automate and optimize business processes. Specifically, we want to compose an existing process
with ``AI skills'' or services so as to generate an optimized 
process that is automated to the extent possible.
The motivation for this comes directly from an extensive body of work 
under the umbrella of ``web service composition'' \cite{papazoglou2003service}.

\subsubsection{Web Service Composition}

The problem of composing web services finds a 
ready ally \cite{dong2004similarity,araghi2012customizing} 
in automated planning techniques since the latter inherently deals 
with the task of composing together actions
in the service of constructing a
course of action or a {\em plan}.
There exists a rich variety of planning techniques, each with 
their own assumptions and features, and it follows
that composition techniques built on top of them 
also demonstrate properties that can be traded of
depending on the needs of the deployment.
We refer to \cite{rao2004survey} for a great summary of work done in this area, 
while \cite{srivastava2003web,sohrabi2010customizing} provides a very useful summary of many of the challenges involved.

\subsection{Competing Philosophies: Head to Head}

We begin with a very brief survey (to the extent
possible within the confines of a conference paper) of existing 
composition techniques and discuss their unique trade-offs.
This will aid in uniquely placing our tool among existing
techniques for web service composition -- and by extension
the optimization of business processes using web services -- 
for clients trying to choose between competing technologies.

\subsubsection{Classical \& Multi-Agent Planning}

Automated planning offer a concise way of describing and maintaining
process -- it is not surprising that the exploration of web service composition
in planning literature begun with classical (deterministic, fully observable) planning or compilations into it \cite{hoffmann2007web}.
From the planning point of view, sophisticated interaction between services 
can be dealt with at the level of reasoning with capabilities of multi-agent
systems \cite{au2008planning}.
A typical feature of web services is the uncertain nature of 
their execution \cite{carman2003web}
which can be modeled in classical form and dealt with at execution time in
certain circumstances \cite{pistore2005automated}.

\subsubsection{Non-deterministic / Probabilistic Planning}

A more natural way of dealing with uncertainty 
is to reason about it during planning itself. 
This can be done offline, by planning with a probability distribution 
\cite{hoffmann2006conformant} or set \cite{muise2012improved}
of possible outcomes. 
As an example, consider calling an API that returns employee
information of a company. 
The call may return the information or hit a 404 error.
A non-deterministic planner will thus plan for both contingencies 
instead of waiting till execution time to reason with the outcome.
Of course, a critical assumption here is that nothing changes 
between planning to execution \cite{sohrabi2010preference}.
Furthermore, since the planning is done offline, dealing with
uncertainty in an open world becomes tricky.

\subsubsection{Robust (Model-Lite) Planning}

Theoretically, every domain is deterministic 
if you can model all the variables. 
However, a complete model is rarely available, 
and incompleteness can manifest itself in a different form 
of uncertainty at the time of planning.
Let's go back to the example of the API that returns 
employee information. 
Imagine that the API can only be pinged from inside 
the company firewall, but this constraint is not 
part of the planner's model.
This uncertainty will be resolved at the time of execution
(since the domain is deterministic, you will get the same outcome
every time you ping it) but the planner can account for 
this uncertainty at planning time. 
This flavor or planning, referred to as ``model-lite planning'' \cite{kambhampati2007model} was proposed especially keeping web service composition in mind. 
Planning in this paradigm tries to maximize success in the most number
of possible models, i.e. the {\em robustness} \cite{nguyen2017robust} of a plan.
In our running example, the planner would string together
API calls that can potentially provide the same information in order
to maximize chances of success (instead of contingent
solutions like non-deterministic planners).
We will see later, given the size of our compositions, 
stringing together so many services is probably not a viable solution.

\subsubsection{Replanning}

Replanning offers a viable alternative to planning with
uncertain outcomes upfront \cite{little2007probabilistic}. 
The replanning strategy, of course, varies largely based on 
assumptions about the underlying domain as we discussed above \cite{yoon2007ff}
-- e.g. in the context of robust planning, doing the same action multiple times
will yield the same result since the underlying domain is deterministic.
The replanning strategy also depends on the what properties of the new 
plan one wants to optimize -- e.g. whether it is desired that the new 
process is as close to the older one as possible or it preserves 
key properties or commitments \cite{talamadupula2013theory}.

Replanning significantly increases the runtime complexity, 
but is necessary at the end of the day especially since most models
are not complete and will eventually require replanning even
with the most carefully constructed plan dealing with possible
contingencies. In the context of business processes, for example, 
replanning has been used in the past for the adaptation
task \cite{bucchiarone2011adaptation}.

\subsubsection{Hierarchical Planning}

Hierarchical Task Networks or HTNs based planners such as SHOP2
\cite{nau2003shop2} 
provide a powerful and alternative framework for composition of services.
This includes unique features such as baking in considerations 
of quality of service into the action theory 
and the ability to sort preconditions to effect this. 
The HTN framework has seen continued interest in the space of 
web service composition tasks \cite{sirin2004htn} outside the scope of
more traditional planning mechanisms.

\subsubsection{Action Languages}

Action languages provide another interesting 
alternative -- in \cite{sohrabi2009web} authors
used Golog-based templating to represent preferences, while
authors in \cite{marrella2014smartpm-1,marrella2014smartp-2} use it 
to address the adaptation task.

\begin{figure*}
\centering
\includegraphics[width=\textwidth]{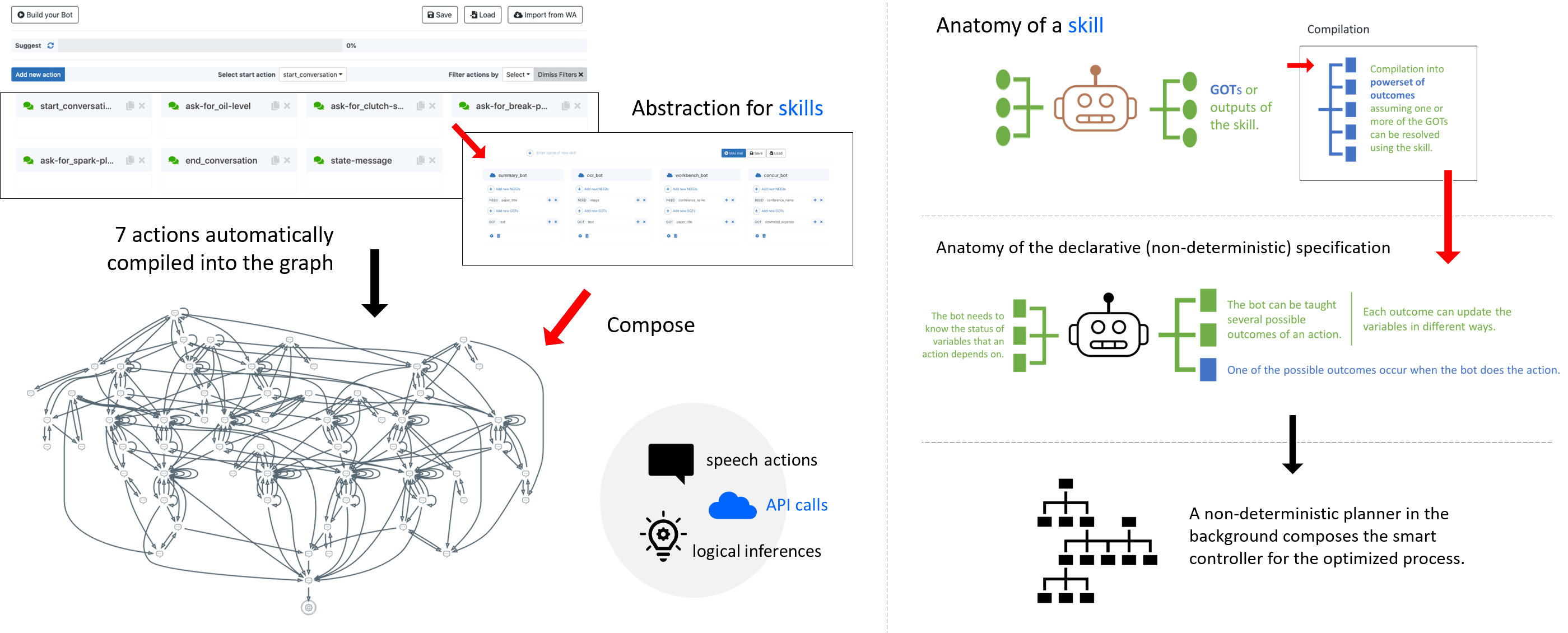}
\caption{
Overview of the flavor of declarative specification adopted in \dwa and the construction of \dba on top of it.
Parts of this image has been reproduced with permission from \cite{muise2019planning}.
{\bf Demo link}: \textcolor{blue}{\url{https://bit.ly/33RNMiR}}.
}
\label{fig:d3wa}
\end{figure*}

\subsection{\dba Value Proposition}

Our tool \dba brings the notion of web service composition
to the task of business process management 
\cite{marrella2017automated}.
The idea here is to be able to change, evolve, and optimize 
a business process by injecting into it services
that can perform specific computation, while at the same 
time providing the persona in charge of the process
tools to manage it easily. 
The specific notion of optimization we focus on
is that of maximizing automation in the composed
process and thus reduce the load
on individual case workers. Of course, if the
automated components come with additional features
we can add those considerations (such as health, 
probability of success, cost, etc.) into the
optimization criterion. 

Thus \dba comes with two main features:

\begin{itemize}
\item[(1)] an interface to build, edit, visualize, and maintain 
complex processes concisely using a declarative specification 
that allows exponential scale-up from the representation to the
realized process; and
\item[(2)] an interface to optimize the process by composing it with services that can automated one or more parts of it while still maintaining the features from (1).
\end{itemize}

In order to achieve this,
we build on a substrate of non-deterministic planning 
so as to be able to {\bf (a)} compute the composed process
offline and allow the process manager to analyse 
and edit it; while also {\bf (b)} be able to plan for
the inherent uncertain nature of execution with 
external services.
The tool also comes with its own inbuilt executor 
for the generated process once it is deployed.
This can be used to optimize a part of a given
business process or orchestrate the entire 
life-cycle of the process.

\begin{figure*}
\centering
\begin{subfigure}[b]{0.55\textwidth}
\includegraphics[width=\textwidth]{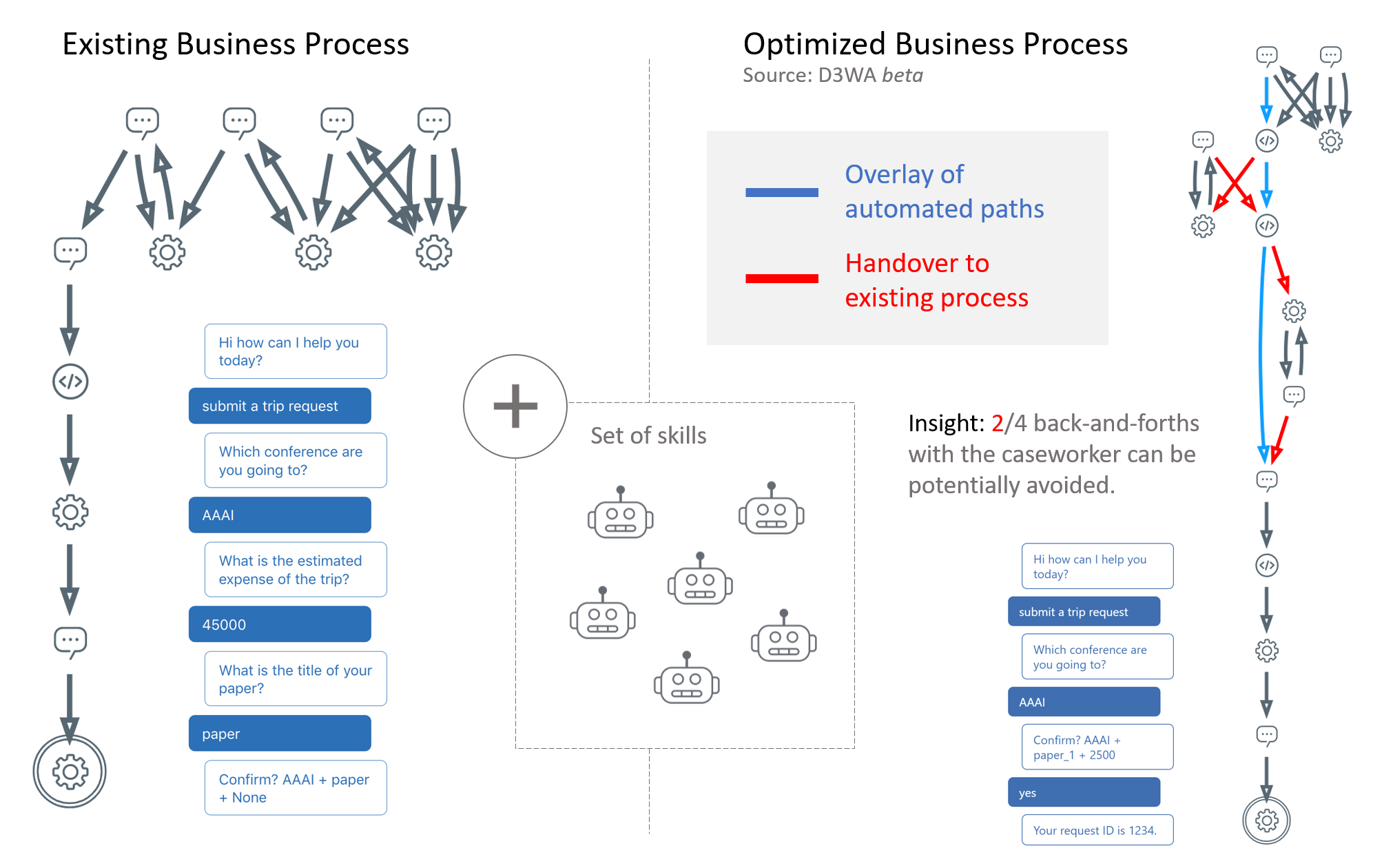}
\caption{Composition illustrating fallbacks to original process.}
\label{fig:fallbacks}
\end{subfigure}
\hfill
\begin{subfigure}[b]{0.15\textwidth}
\includegraphics[width=\textwidth]{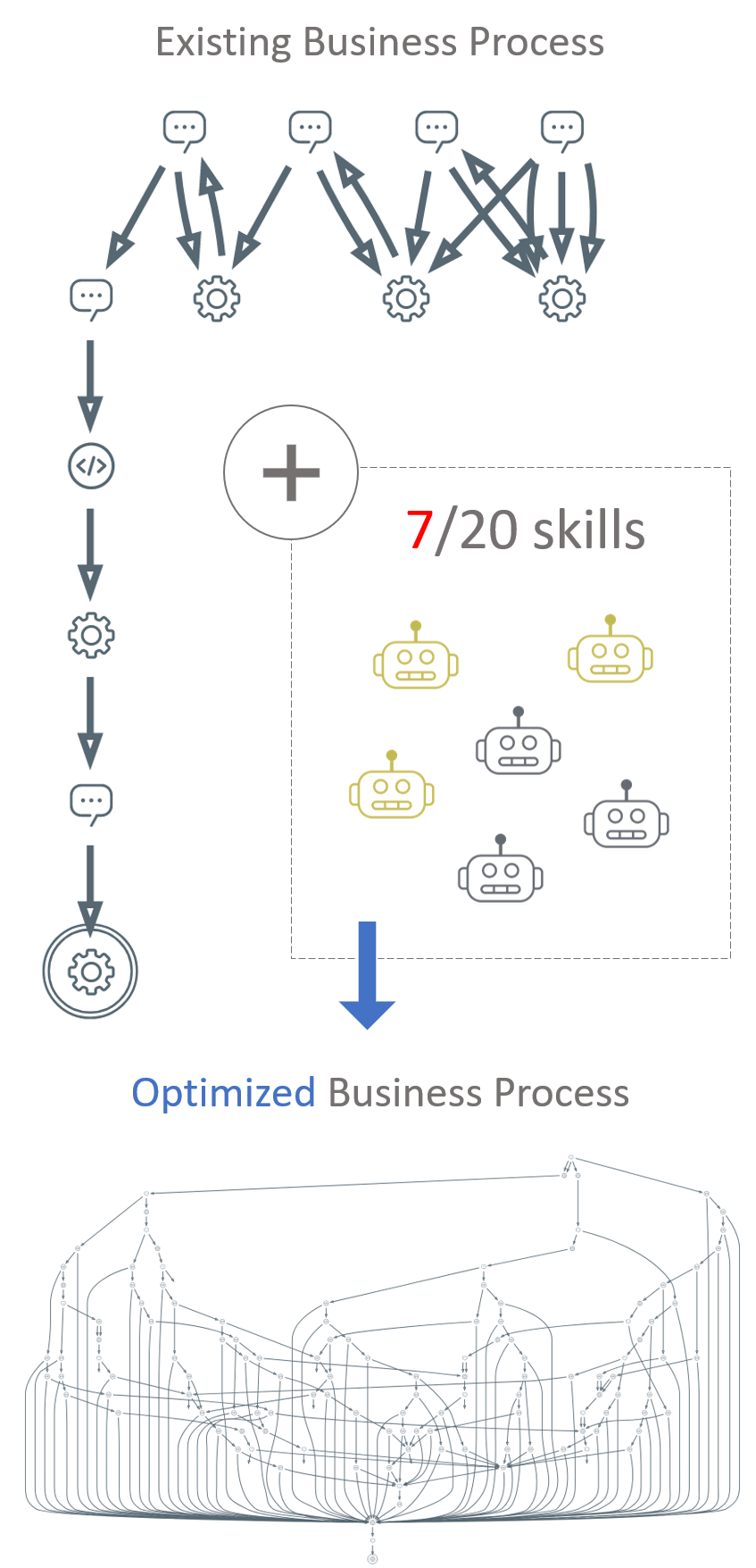}
\caption{Robustness.}
\label{fig:robustness}
\end{subfigure}
\hfill
\begin{subfigure}[b]{0.17\textwidth}
\includegraphics[width=\textwidth]{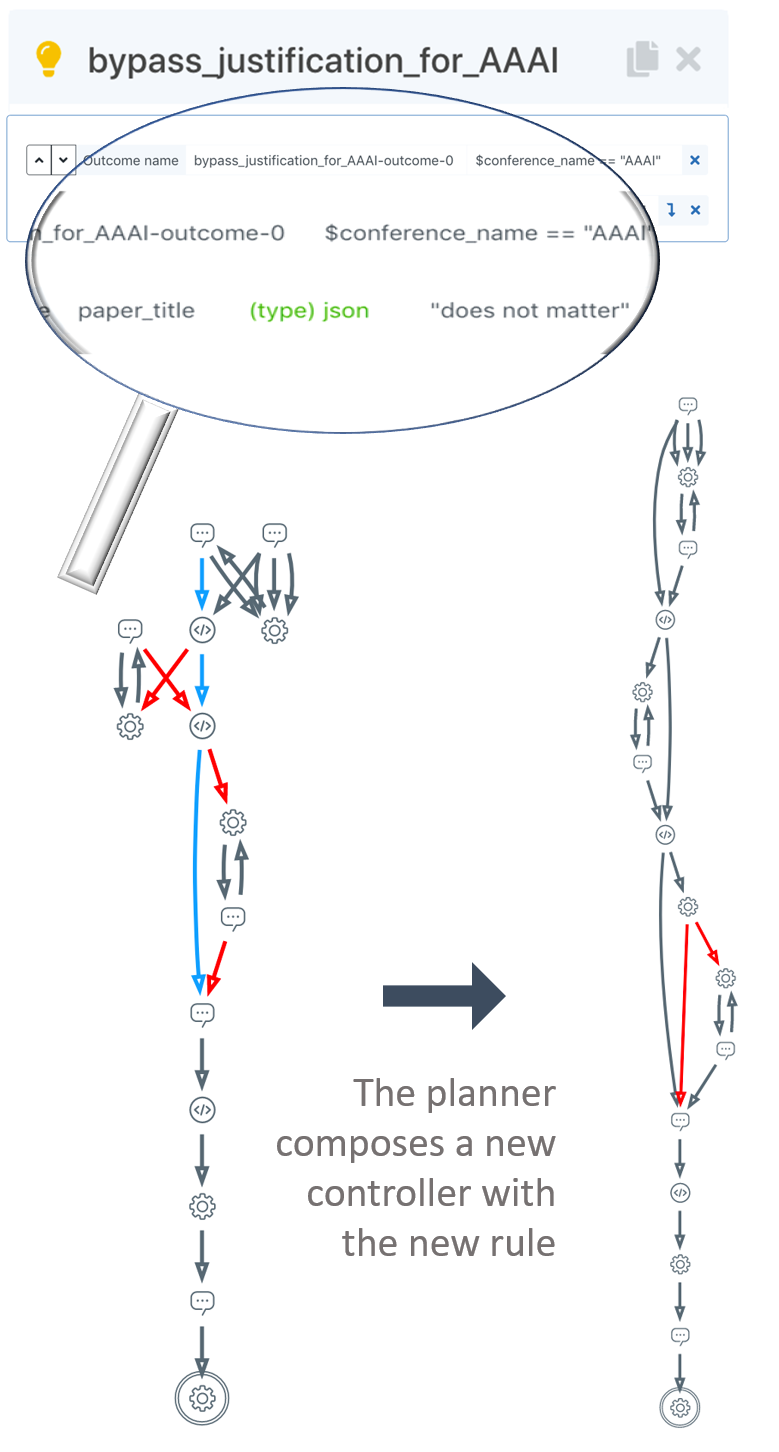}
\caption{Personalization.}
\label{fig:personalization}
\end{subfigure}
\caption{Examples illustrating salient properties of \dba compositions,
including
(\ref{fig:fallbacks}) fallback to equivalent services or (in the worst case) 
to the original manual process;
(\ref{fig:robustness}) automated filtering of relevant services + chaining of equivalent skills to increase robustness of the composition; and finally
(\ref{fig:personalization}) easy personalization and management of the composed process.}
\label{fig:animals}
\end{figure*}

With regard to past work at the intersection of non-deterministic planning 
and declarative specification of processes, we would like to emphasize that our focus here is not on the declarative aspect as a means to define a process but as a means to facilitate a particular kind of optimization (through composition with automation) and also not on non-determinism as a means to handle uncertainty (which comes for free) but as a means to facilitate the user interaction for this optimization process. There is indeed a lot of work \cite{pesic2006declarative,srivastava2004decision}
on how the declarative paradigm readily translates to business process applications that require the definition and composition of processes. That particular aspect of the problem is admittedly quite well explored but only acts as a means to an end here. 

The unique value proposition we are providing is the 
{\em business process + skills = optimized business process interaction patterns} with the human in the loop at design time \cite{marrella2017planning}. The proposed interaction is a very specific flavor of human-in-the-loop composition that allows the business process manager persona to take in a new process or write a process in a declarative form, take in a catalog of skills, and author optimized processes offline. This is where the proposed framework built on non-deterministic planning comes into form in being able to surface different optimized compositions to the manager persona to edit, debug, visualize, and personalize further. The declarative form is necessary for this to happen (as well as account for the fact that skill catalogs are independent of the process).

\section{Background: \dwa}

In AAAI 2019, \citeauthor{mai} demonstrated \dwa\ 
-- a tool meant to bring down the effort and expertise
required to design sophisticated {\em goal-directed} conversational agents e.g. for applications such as customer support.
The state of the art \cite{kinaSUR} in the design of such agents requires the dialogue designer to either write down
the entire dialogue tree by hand (e.g. Google Dialogue Flow
or Watson Assistant) which becomes intractable pretty soon
or train end-to-end systems which 
provide no control over their emerging behavior 
\cite{tay} and are thus unusable in the enterprise scene.
Instead, in \cite{muise2019planning,botea2019generating}, the authors proposed a paradigm shift in how such agents are built by conceiving a declarative way of specifying them. 
In this paper, we extend their framework for the purposes of the definition and composition of automated skills into business process specifications.

\subsection{Anatomy of a Declarative Specification} 

At the core of the declarative specification
(as shown in Figure~\ref{fig:d3wa}
is a set of variables that model the state of the world
and actions that depend and operate on those variables
to define capabilities that an agent has to affect 
change to the world.
In the context of a conversational agent, 
such actions can be either speech actions that interact
directly with the end user, or internal actions
such as logical inferences or API calls. 
The latter, of course, is more relevant to our case.

Each action is defined by a set of NEEDs or statuses of variables
which tells the agent when it can perform the action (these become preconditions of actions available to the planner in the backend), 
and a set of OUTCOMES (recall the two outcomes in the example of an API call from before), one of which might occur at execution time. 
Each outcome produces a set of updates to the variables
-- these are compiled to the non-deterministic effects of the 
actions available to the planner in the backend.
A non-deterministic planner\footnote{
We use {\tt PRP} \cite{muise2012improved} 
as the non-deterministic planner 
and Hovor \cite{fss}
as the execution engine.
The intent classifiers that make up the determiners 
for the action outcomes are based on 
Watson Assistant {\em Beta} features.} 
in the backend uses this specification 
to plan all possible outcomes offline and automatically generates the 
resulting dialogue tree (which would otherwise have had to be
written manually).
This results in an exponential savings from the size of the 
specification to the complexity of the composed agent,
as shown in Figure~\ref{fig:d3wa}.
For more details on \dwa we refer the reader to \cite{muise2019planning}.
In the following section, we will describe how we adopt
this platform for the management and optimization of 
business processes.

\section{\dba: Process + Skills = Optimized Process}

This section outlines the basics of the proposed composition technique 
with illustrative examples. A video demonstration can be viewed at 
\textcolor{blue}{\url{https://bit.ly/33RNMiR}}.

\subsubsection{Business Processes}

The starting point of the composition process is a business
process written either inside \dba itself, or 
an existing process imported 
from an outside source 
(e.g. as in Figure~\ref{fig:travel}), 
usually in the form of a finite state machine, 
a graph, a mind map, or a similar data structure. 
Though the \dba tool currently does not support the latter yet, 
refer to \cite{sohrabi2018ai} how this translation is done
(note that this translation will lose the exponential savings 
that can be gained from a declarative specification).

The interface to specifying a process remains almost identical
to \dwa. This is not surprising since a key aspect of 
designing goal-oriented dialogue is the specification of 
the underlying process that must be maintained in
conversation (e.g. in customer support).
The only difference is that certain types of actions,
namely the API calls and the logic actions, take more
precedence over the speech actions. 
The latter may or may not occur in a process at all 
but we keep them around to facilitate a conversational
interface to the processes (as we will see later).
These form a sufficient set of capabilities 
to represent any business process. 

\subsubsection{Skills}

As we mentioned before, 
our goal is to augment existing business processes 
with automated components, so as to optimize the overall task. 
We call these automated components: {\em skills}.
This notion of skills is consistent with skills in 
Watson Assistant\footnote{Watson Assistant Skills:  \textcolor{blue}{\url{https://ibm.co/2LblJ70}}} 
or Amazon Alexa \footnote{Amazon Alexa Skills: \textcolor{blue}{\url{https://amzn.to/2ZH9Olp}}}
as a function that can perform microtasks.
We assume a {\em catalog of skills}\footnote{
Once a new process has been composed,
it can be packaged into a new service and folded
into the skill catalog. 
Such ``complex actions'' have been explored
before in \cite{mci-fad-nmr02f}.
} accessible to the persona
designing and managing the process, which she 
can import and hit compose on the interface.

The skill specification interface follows a similar structure 
to that of a generic \dwa action
but provides an even more simpler abstraction to the outcome enumeration. 
It accepts as NEEDs and GOTs the inputs and outputs of a service. 
The declarative specification of these skills, to be consumed alongside that of the rest of the business process, is compiled internally from this abstract specification by considering a power set of the GOTs as the possible OUTCOMEs
of invoking the service. 
The semantics of this compilation is that by invoking the service
the planner is expecting to get back (n)one or more of the promised
outputs.

\subsubsection{Composition Technique}

The inputs to the composition step is thus a process
and a set of skills and the output is a optimized process 
wherein the original process has been composed with skills 
wherever possible to maximize automation, as
shown in Figure \ref{fig:animals}.
Once the skills have been compiled to the standard \dwa form
the rest of the process remains same as in \dwa.
This means we get all the rest of its features for free,
including being able to visualize, debug, and iterate on the 
composed process once it has been computed. 
Specifically with regards to business process management,
we illustrate some key capabilities next.

The reason declarative works well in this setting is two-fold: 
First, the sheer size of these composed processes, and the need to be able to be flexible with their management, makes it imperative that they are not written and maintained by hand. Furthermore, as we mentioned before, the source of skills and processes may be different. The declarative framework allows developers of either to develop without having to worry about how they relate to each other. The planner preforms an essential role in the background 
by providing a powerful tool to stitch them together.

It is important to note that this composition task is quite non-trivial. 
This is because, given a complex business process and a large set of skills, it is computationally intractable by a human being to figure out all the possible combinations and find the optimal ones among them. 
Further, the source of the process and the skills may be different. For example, a developer who is writing the skill may have no idea about the business processes that their skill is eventually going to be used in. 
Furthermore, the features of these skills, as well as the process itself, change over time, making it essential that the composition process is automated. 

\subsection{Features of Proposed Composition}

Consider the travel application process again (see Figure~\ref{fig:animals}).
On the left, we see a part of the travel approval business process
dealing with acquiring information from the applicant. 
There are three back-and-forths with the employee, to figure out the name of the conference, title of the presentation, and estimated expenses, until this
step of the process is complete. 

\subsubsection{Fallbacks}

On the right of Figure~\ref{fig:fallbacks}, we see the optimized business process composed out of skills. These skills, for example, may be able to come up with estimated expenses given a conference\footnote{Built on the Amadeus API: \textcolor{blue}{\url{https://developers.amadeus.com/}}.}, or look up the paper title given the employee information\footnote{Built on top of the (IBM internal) Author Workbench service.}. 
In the optimized process, the added flows due to the automated skills are shown in blue, while handovers from automation to the original process is highlighted in red. Notice that the four original back-and-forths are still there, but two of them have been bypassed by the skills and are only resorted to as fallbacks, if they fail. This is noticeable in the dialogue in the inset, where the number of interactions with the caseworker is reduced. 

\subsubsection{Robustness}

These fallbacks need not be restricted to the original process only.
As shown in Figure~\ref{fig:robustness}, 
our approach figures out how to chain equivalent skills 
to maximize success of the automated components.
The composition technique allows this optimized process to grow
exponentially to increase the robustness of automation.
This is an example of how quickly the composition task 
(let alone the orchestration -- i.e. execution and 
monitoring -- of the process once it is composed)
can go out of hand for manual approaches even for 
the small process used here for illustrative purposes. 

\subsubsection{Relevancy}

An additional feature to note in Figure~\ref{fig:robustness} 
is that the catalog is quite larger than the set of skills
finally finding a role in the optimized process. 
Our approach figures out which skills to use (and when and where) and 
which to ignore from the skill catalog. 
In addition to the complexity or chained composition demonstrated above,
this too is often beyond the scope of manual composition.

\subsubsection{Customization}

Finally, our approach also makes it very easy to modify and personalize the composed process with your own rules. 
In Figure~\ref{fig:personalization}, for example, we have added a rule to bypass the manager approval if the presentation is going to happen at a particular conference. 
This is automatically reflected again, in the newly composed process.
This ability to make small edits and effect large changes in the process
is a unique feature of declarative modeling.

\begin{table}[]
\small
\begin{tabular}{@{}rccccc@{}}
\toprule
number of skills        & Base  & 5    & 10   & 15   & 20    \\ \midrule
size of composition (\#edges) & 23 & 134  & 209  & 370  & 1039  \\
time to generate (secs) & -  & 0.02 & 0.24 & 3.10 & 19.48 \\ \bottomrule
\end{tabular}
\caption{Average size of composed processes with respect to number of skills 
over 5 randomly generated catalogs.}
\label{tab:scale}
\end{table}

\begin{figure}
\centering
\includegraphics[width=\columnwidth]{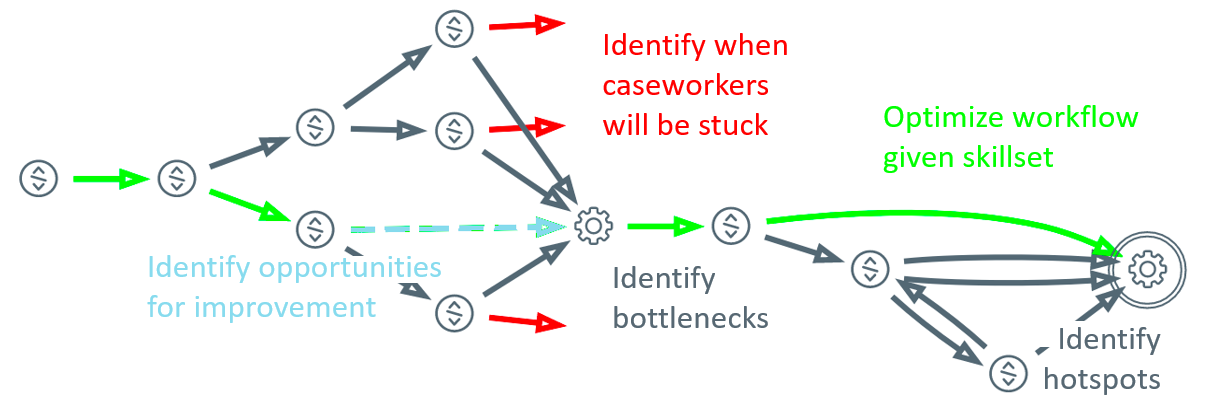}
\caption{Challenge problem: meta-reasoning to improve an existing process by searching in the space of models.}
\label{fig:meta}
\end{figure}

\subsection{Composition Complexity}

Before we end, we would like to give a sense of the complexity 
of the compositions. 
Table \ref{tab:scale} illustrates the sizes (measured in edges to compare against the complexity of manual build) of the composed 
processes and runtime with respect to increasing size of the skill catalog.
Clearly, while composition is intractable without automation even for 
small catalog sizes, the planner is quite fast (especially considering 
the compositions are offline) in coming up with the solutions.

\section{Challenge Problems}

We end with a discussion of two key research challenges 
that come out of the \dba paradigm.

\subsubsection{Meta-Reasoning on Business Processes}

So far we have concentrated only 
on improving existing processes using skills.
The improvements have been closed world, 
i.e. confined to the existing process,
and underlying process or model never changes. 
The promise of \dba
need not be confined to this.
In fact, the planning community has recently seen
increased attention to the ability to search and reason 
in the space of models \cite{keren2019efficient,explain}.
Such techniques can allow us to evolve a process 
to maximize measurable performance metrics
such as risk \cite{muthusamy2018towards}.
Recent work has take the first steps \cite{narendra2019counterfactual} 
towards evolving business processes to that effect.
Model space reasoning can be used to identify opportunities for improving a workflow by addressing hot-spots, bottlenecks, and dead-ends in a process (Figure~\ref{fig:meta}).
Works on process mining for planning specifications \cite{de2016computing,de2017disruptive} can also provide powerful 
tools to his end.

\subsubsection{Taxonomy and Knowledge Base}

One of the assumptions we made throughout this work
was that the skill developers and 
the persona in charge of managing the business process, though
working independently, still have access to a shared vocabulary
or taxonomy to specify artifacts of their processes or skills.
Ideally, we would like to relax this and instead be able to call to 
a knowledge base in the background that can make 
a suitable mapping between the two specifications. 
This knowledge can reside in an ontology in the backend 
or may, for example, come from meta information provided by the 
skill developer in the manifest file corresponding to their skill.
Existing approaches such as Embedded Business AI Framework\footnote{Embedded Business AI Framework: \textcolor{blue}{\url{https://www.eba.ai/}}} (EBA) has attempted to provide a tight integration with skills and ontologies in the past, and can potentially
provide inspiration towards enabling more sophisticated composition
strategies in \dba as well.
The now sunsetted project API Harmony\footnote{API Harmony: \textcolor{blue}{\url{https://apiharmony-open.mybluemix.net}}} 
can also provide valuable lessons on the representation and 
composition of diverse web services and skills.
In the context of planning in particular, authors in \cite{hoffmann2007web}
have previously attempted to leverage ontologies in the context of web service composition, while works in business process management \cite{hepp2005semantic} has also proposed similar solutions for this problem.

\section{Looking Forward: Path to Deployment}

In this paper, we demonstrated how non-deterministic planning can 
provide a powerful way of constructing, maintaining,
and optimizing business processes using AI microservices.
Going forward, we will be looking to integrate this tooling
into IBM's Digital Business Autamation\footnote{DBA: \textcolor{blue}{\url{https://www.ibm.com/cloud/garage/architectures/dba}}} (DBA) 
suite aimed to streamline business operations through automation.
The larger effort towards this integration can be read in \cite{yara}.

\subsection{Acknowledgements} 
We thank IBM's digital business automation team and the research team 
including Yara Rizk, Scott Boag, Vatche Isahagian, Falk Pollok, Vinod Muthusamy, Sampath Dechu, Merve Unuvar, and Rania Khalaf for their support and ideas throughout the project. 
A special word of thanks to Christian Muise and the rest of the \dwa team \cite{muise2019planning}
on whose work we built our \dba extension.
Finally, many thanks to Shirin Sohrabi and Michael Katz, also from
IBM Research, who helped us navigate the fascinating world 
of planning, business processes, and web service composition. 

\bibliographystyle{aaai}
\bibliography{bib}

\begin{thebibliography}{}

\bibitem[\protect\citeauthoryear{Araghi}{2012}]{araghi2012customizing}
Araghi, S.~S.
\newblock 2012.
\newblock {\em Customizing the Composition of Web Services and Beyond}.
\newblock Ph.D. Dissertation, U Toronto.

\bibitem[\protect\citeauthoryear{Au, Kuter, and Nau}{2008}]{au2008planning}
Au, T.-C.; Kuter, U.; and Nau, D.
\newblock 2008.
\newblock {Planning for Interactions Among Autonomous Agents}.
\newblock In {\em International Workshop on Programming Multi-Agent Systems}.

\bibitem[\protect\citeauthoryear{Botea \bgroup et al\mbox.\egroup
  }{2019}]{botea2019generating}
Botea, A.; Muise, C.; Agarwal, S.; Alkan, O.; Bajgar, O.; Daly, E.; Kishimoto,
  A.; Lastras, L.; Marinescu, R.; Ondrej, J.; Pedemonte, P.; and Vodolan, M.
\newblock 2019.
\newblock {Generating Dialogue Agents via Automated Planning}.
\newblock {\em arXiv:1902.00771}.

\bibitem[\protect\citeauthoryear{Bucchiarone \bgroup et al\mbox.\egroup
  }{2011}]{bucchiarone2011adaptation}
Bucchiarone, A.; Pistore, M.; Raik, H.; and Kazhamiakin, R.
\newblock 2011.
\newblock {Adaptation of Service-Based Business Processes by Context-Aware
  Replanning}.
\newblock In {\em SOCA}.

\bibitem[\protect\citeauthoryear{Carman, Serafini, and
  Traverso}{2003}]{carman2003web}
Carman, M.; Serafini, L.; and Traverso, P.
\newblock 2003.
\newblock {Web Service Composition as Planning}.
\newblock In {\em ICAPS Workshop on Planning for Web Services}.

\bibitem[\protect\citeauthoryear{Chakraborti \bgroup et al\mbox.\egroup
  }{2017}]{explain}
Chakraborti, T.; Sreedharan, S.; Zhang, Y.; and Kambhampati, S.
\newblock 2017.
\newblock {Plan Explanations as Model Reconciliation: Moving Beyond Explanation
  as Soliloquy}.
\newblock In {\em IJCAI}.

\bibitem[\protect\citeauthoryear{Chakraborti \bgroup et al\mbox.\egroup
  }{2019}]{mai}
Chakraborti, T.; Muise, C.; Agarwal, S.; and Lastras, L.
\newblock 2019.
\newblock {D3WA : An Intelligent Model Acquisition Interface for Interactive
  Specification of Dialog Agents}.
\newblock In {\em AAAI \& ICAPS Demo Tracks}.

\bibitem[\protect\citeauthoryear{De~Giacomo \bgroup et al\mbox.\egroup
  }{2016}]{de2016computing}
De~Giacomo, G.; Maggi, F.~M.; Marrella, A.; and Sardina, S.
\newblock 2016.
\newblock Computing trace alignment against declarative process models through
  planning.
\newblock In {\em ICAPS}.

\bibitem[\protect\citeauthoryear{De~Giacomo \bgroup et al\mbox.\egroup
  }{2017}]{de2017disruptive}
De~Giacomo, G.; Maggi, F.~M.; Marrella, A.; and Patrizi, F.
\newblock 2017.
\newblock On the disruptive effectiveness of automated planning for ltl f-based
  trace alignment.
\newblock In {\em AAAI}.

\bibitem[\protect\citeauthoryear{de Leoni, Lanciano, and
  Marrella}{2018}]{de2018aligning}
de~Leoni, M.; Lanciano, G.; and Marrella, A.
\newblock 2018.
\newblock {Aligning Partially-Ordered Process-Execution Traces and Models Using
  Automated Planning}.
\newblock In {\em ICAPS}.

\bibitem[\protect\citeauthoryear{Dong \bgroup et al\mbox.\egroup
  }{2004}]{dong2004similarity}
Dong, X.; Halevy, A.; Madhavan, J.; Nemes, E.; and Zhang, J.
\newblock 2004.
\newblock {Similarity Search for Web Services}.
\newblock In {\em VLDB}.

\bibitem[\protect\citeauthoryear{Hepp \bgroup et al\mbox.\egroup
  }{2005}]{hepp2005semantic}
Hepp, M.; Leymann, F.; Domingue, J.; Wahler, A.; and Fensel, D.
\newblock 2005.
\newblock {Semantic Business Process Management: A Vision Towards Using
  Semantic Web Services for Business Process Management}.
\newblock In {\em ICEBE}.

\bibitem[\protect\citeauthoryear{Hoffmann and
  Brafman}{2006}]{hoffmann2006conformant}
Hoffmann, J., and Brafman, R.~I.
\newblock 2006.
\newblock {Conformant Planning via Heuristic Forward Search: A New Approach}.
\newblock {\em AIJ}.

\bibitem[\protect\citeauthoryear{Hoffmann, Bertoli, and
  Pistore}{2007}]{hoffmann2007web}
Hoffmann, J.; Bertoli, P.; and Pistore, M.
\newblock 2007.
\newblock {Web Service Composition as Planning, Revisited: In Between
  Background Theories and Initial State Uncertainty}.
\newblock In {\em AAAI}.

\bibitem[\protect\citeauthoryear{Hull and Nezhad}{2016}]{hull2016rethinking}
Hull, R., and Nezhad, H. R.~M.
\newblock 2016.
\newblock {Rethinking BPM in a Cognitive World: Transforming How We Learn and
  Perform Business Processes}.
\newblock In {\em BPM}.

\bibitem[\protect\citeauthoryear{Jarvis \bgroup et al\mbox.\egroup
  }{1999}]{jarvis1999exploiting}
Jarvis, P.; Moore, J.; Stader, J.; Macintosh, A.; Casson-du Mont, A.; and
  Chung, P.
\newblock 1999.
\newblock {Exploiting AI Technologies to Realise Adaptive Workflow Systems}.
\newblock In {\em AAAI Workshop on Agent-Based Systems in the Business
  Context}.

\bibitem[\protect\citeauthoryear{Kambhampati}{2007}]{kambhampati2007model}
Kambhampati, S.
\newblock 2007.
\newblock {Model-Lite Planning for the Web Age Masses: The Challenges of
  Planning with Incomplete and Evolving Domain Models}.
\newblock In {\em AAAI}.

\bibitem[\protect\citeauthoryear{Keren \bgroup et al\mbox.\egroup
  }{2019}]{keren2019efficient}
Keren, S.; Pineda, L.; Gal, A.; Karpas, E.; and Zilberstein, S.
\newblock 2019.
\newblock {Efficient Heuristic Search for Optimal Environment Redesign}.
\newblock In {\em ICAPS}.

\bibitem[\protect\citeauthoryear{Little and
  Thiebaux}{2007}]{little2007probabilistic}
Little, I., and Thiebaux, S.
\newblock 2007.
\newblock {Probabilistic Planning vs Replanning}.
\newblock In {\em ICAPS Workshop on IPC: Past, Present and Future}.

\bibitem[\protect\citeauthoryear{Marrella and
  Lesp{\'e}rance}{2017}]{marrella2017planning}
Marrella, A., and Lesp{\'e}rance, Y.
\newblock 2017.
\newblock {A Planning Approach to the Automated Synthesis of Template-Based
  Process Models}.
\newblock {\em Service Oriented Computing and Applications}.

\bibitem[\protect\citeauthoryear{Marrella, Mecella, and
  Sardina}{2017}]{marrella2014smartpm-1}
Marrella, A.; Mecella, M.; and Sardina, S.
\newblock 2017.
\newblock {Intelligent Process Adaptation in the SmartPM System}.
\newblock {\em TIST}.

\bibitem[\protect\citeauthoryear{Marrella, Mecella, and
  Sardina}{2018}]{marrella2014smartp-2}
Marrella, A.; Mecella, M.; and Sardina, S.
\newblock 2018.
\newblock {Supporting Adaptiveness of Cyber-Physical Processes through
  Action-based Formalisms}.
\newblock {\em AI Communications}.

\bibitem[\protect\citeauthoryear{Marrella}{2017}]{marrella2017automated}
Marrella, A.
\newblock 2017.
\newblock {Automated Planning for Business Process Management}.
\newblock {\em Journal on Data Semantics}.

\bibitem[\protect\citeauthoryear{McIlraith and Fadel}{2002}]{mci-fad-nmr02f}
McIlraith, S., and Fadel, R.
\newblock 2002.
\newblock {Planning with Complex Actions}.
\newblock In {\em Workshop on Non-Monotonic Reasoning}.

\bibitem[\protect\citeauthoryear{Metz}{2018}]{tay}
Metz, R.
\newblock 2018.
\newblock {Microsoft's Neo-Nazi Sexbot was a Great Lesson for Makers of AI
  Assistants}.
\newblock MIT Tech. Review.

\bibitem[\protect\citeauthoryear{Muise \bgroup et al\mbox.\egroup }{2019}]{fss}
Muise, C.; Vodolan, M.; Agarwal, S.; Bajgar, O.; and Lastras, L.
\newblock 2019.
\newblock {Executing Contingent Plans: Challenges in Deploying Artificial
  Agents}.
\newblock In {\em AAAI Fall Symposium}.

\bibitem[\protect\citeauthoryear{Muise \bgroup et al\mbox.\egroup
  }{2020}]{muise2019planning}
Muise, C.; Chakraborti, T.; Agarwal, S.; Bajgar, O.; Chaudhary, A.;
  Lastras-Montano, L.~A.; Ondrej, J.; Vodolan, M.; and Wiecha, C.
\newblock 2020.
\newblock {Planning for Goal-Oriented Dialogue Systems}.
\newblock {\em AAAI Workshop on Interactive and Conversational Recommendation
  Systems}.

\bibitem[\protect\citeauthoryear{Muise, McIlraith, and
  Beck}{2012}]{muise2012improved}
Muise, C.; McIlraith, S.; and Beck, C.
\newblock 2012.
\newblock {Improved Non-Deterministic Planning by Exploiting State Relevance}.
\newblock In {\em ICAPS}.

\bibitem[\protect\citeauthoryear{Muthusamy, Slominski, and
  Ishakian}{2018}]{muthusamy2018towards}
Muthusamy, V.; Slominski, A.; and Ishakian, V.
\newblock 2018.
\newblock Towards enterprise-ready ai deployments minimizing the risk of
  consuming ai models in business applications.
\newblock In {\em AI4I}.

\bibitem[\protect\citeauthoryear{Narendra \bgroup et al\mbox.\egroup
  }{2019}]{narendra2019counterfactual}
Narendra, T.; Agarwal, P.; Gupta, M.; and Dechu, S.
\newblock 2019.
\newblock Counterfactual reasoning for process optimization using structural
  causal models.
\newblock In {\em BPM}.

\bibitem[\protect\citeauthoryear{Nau \bgroup et al\mbox.\egroup
  }{2003}]{nau2003shop2}
Nau, D.~S.; Au, T.-C.; Ilghami, O.; Kuter, U.; Murdock, J.~W.; Wu, D.; and
  Yaman, F.
\newblock 2003.
\newblock {SHOP2: An HTN Planning System}.
\newblock {\em JAIR}.

\bibitem[\protect\citeauthoryear{Nguyen, Sreedharan, and
  Kambhampati}{2017}]{nguyen2017robust}
Nguyen, T.; Sreedharan, S.; and Kambhampati, S.
\newblock 2017.
\newblock {Robust Planning with Incomplete Domain Models}.
\newblock {\em AIJ}.

\bibitem[\protect\citeauthoryear{Papazoglou and
  Georgakopoulos}{2003}]{papazoglou2003service}
Papazoglou, M.~P., and Georgakopoulos, D.
\newblock 2003.
\newblock {Service-Oriented Computing}.
\newblock {\em Communications of the ACM}.

\bibitem[\protect\citeauthoryear{Pesic and Van~der
  Aalst}{2006}]{pesic2006declarative}
Pesic, M., and Van~der Aalst, W.~M.
\newblock 2006.
\newblock {A Declarative Approach for Flexible Business Processes Management}.
\newblock In {\em BPM}.

\bibitem[\protect\citeauthoryear{Pistore, Traverso, and
  Bertoli}{2005}]{pistore2005automated}
Pistore, M.; Traverso, P.; and Bertoli, P.
\newblock 2005.
\newblock {Automated Composition of Web Services by Planning in Asynchronous
  Domains}.
\newblock In {\em ICAPS}.

\bibitem[\protect\citeauthoryear{R-moreno \bgroup et al\mbox.\egroup
  }{2007}]{r2007integrating}
R-moreno, M.~D.; Borrajo, D.; Cesta, A.; and Oddi, A.
\newblock 2007.
\newblock {Integrating Planning and Scheduling in Workflow Domains}.
\newblock {\em Expert Systems with Applications}.

\bibitem[\protect\citeauthoryear{Rao and Su}{2004}]{rao2004survey}
Rao, J., and Su, X.
\newblock 2004.
\newblock {A Survey of Automated Web Service Composition Methods}.
\newblock In {\em Workshop on Semantic Web Services and Web Process
  Composition}.

\bibitem[\protect\citeauthoryear{Rizk \bgroup et al\mbox.\egroup }{2017}]{yara}
Rizk, Y.; Bhandwalder, A.; Boag, S.; Chakraborti, T.; Isahagian, V.; Khazaeni,
  Y.; Pollok, F.; and Unuvar, M.
\newblock 2017.
\newblock A unified conversational assistant framework for business process
  automation.
\newblock In {\em AAAI Worskhop on Intelligent Process Automation (IPA)}.

\bibitem[\protect\citeauthoryear{Sirin \bgroup et al\mbox.\egroup
  }{2004}]{sirin2004htn}
Sirin, E.; Parsia, B.; Wu, D.; Hendler, J.; and Nau, D.
\newblock 2004.
\newblock {HTN Planning for Web Service Composition Using SHOP2}.
\newblock {\em Web Semantics: Science, Services and Agents on the World Wide
  Web}.

\bibitem[\protect\citeauthoryear{Sohrabi and
  McIlraith}{2010}]{sohrabi2010preference}
Sohrabi, S., and McIlraith, S.~A.
\newblock 2010.
\newblock {Preference-Based Web Service Composition: A Middle Ground Between
  Execution and Search}.
\newblock In {\em ISWC}.

\bibitem[\protect\citeauthoryear{Sohrabi \bgroup et al\mbox.\egroup
  }{2018}]{sohrabi2018ai}
Sohrabi, S.; Riabov, A.~V.; Katz, M.; and Udrea, O.
\newblock 2018.
\newblock {An AI Planning Solution to Scenario Generation for Enterprise Risk
  Management}.
\newblock In {\em AAAI}.

\bibitem[\protect\citeauthoryear{Sohrabi, Prokoshyna, and
  McIlraith}{2009}]{sohrabi2009web}
Sohrabi, S.; Prokoshyna, N.; and McIlraith, S.~A.
\newblock 2009.
\newblock {Web Service Composition via the Customization of Golog Programs with
  User Preferences}.
\newblock In {\em Conceptual Modeling}.

\bibitem[\protect\citeauthoryear{Sohrabi}{2010}]{sohrabi2010customizing}
Sohrabi, S.
\newblock 2010.
\newblock {Customizing the Composition of Actions, Programs, and Web Services
  with User Preferences}.
\newblock In {\em ISWC}.

\bibitem[\protect\citeauthoryear{Sreedhar}{2018}]{kinaSUR}
Sreedhar, K.
\newblock 2018.
\newblock {What it Takes to Build Enterprise-Class Chatbots}.
\newblock Chatbots.

\bibitem[\protect\citeauthoryear{Srivastava and
  Koehler}{2003}]{srivastava2003web}
Srivastava, B., and Koehler, J.
\newblock 2003.
\newblock {Web Service Composition -- Current Solutions and Open Problems}.
\newblock In {\em ICAPS Workshop on Planning for Web Services}.

\bibitem[\protect\citeauthoryear{Srivastava}{2004}]{srivastava2004decision}
Srivastava, B.
\newblock 2004.
\newblock {A Decision-Support Framework for Component Reuse and Maintenance in
  Software Project Management}.
\newblock In {\em CSMR}.

\bibitem[\protect\citeauthoryear{Talamadupula \bgroup et al\mbox.\egroup
  }{2013}]{talamadupula2013theory}
Talamadupula, K.; Smith, D.~E.; Cushing, W.; and Kambhampati, S.
\newblock 2013.
\newblock {A Theory of Intra-Agent Replanning}.
\newblock In {\em ICAPS Workshop on Distributed and Multi-Agent Planning}.

\bibitem[\protect\citeauthoryear{Yoon, Fern, and Givan}{2007}]{yoon2007ff}
Yoon, S.~W.; Fern, A.; and Givan, R.
\newblock 2007.
\newblock {FF-Replan: A Baseline for Probabilistic Planning}.
\newblock In {\em ICAPS}.

\end{thebibliography}

\end{document}